\documentclass{article} 
\usepackage{iclr2025_conference,times}

\usepackage{amsmath,amsfonts,bm}

\def\eqref#1{equation~\ref{#1}}

\def\1{\bm{1}}

\DeclareMathAlphabet{\mathsfit}{\encodingdefault}{\sfdefault}{m}{sl}
\SetMathAlphabet{\mathsfit}{bold}{\encodingdefault}{\sfdefault}{bx}{n}

\usepackage{hyperref}
\usepackage{url}
\usepackage{svg}

\title{Multi-turn Training with basic human feedback helps little on LLM reasoning}

 \author{\textbf{Qiang Liu\textsuperscript{1}\thanks{Q.L.Liu@hotmail.com} \ , Wuganjing Song\textsuperscript{1,}\textsuperscript{2}\thanks{wsongan@connect.ust.hk} \ , Zhenzhou Lin\textsuperscript{1}, Feifan Chen\textsuperscript{1,}\textsuperscript{3},}\\
 \textbf{Qiaolong Cai\textsuperscript{1}, Chen Li\textsuperscript{1}, Yongduo Sui\textsuperscript{1}}\\
\textsuperscript{1}Tencent Interactive Entertainment, Shenzhen, China \\
\textsuperscript{2}The Hong Kong University of Science and Technology, Hong Kong SAR, China\\
\textsuperscript{3}Sun Yat-sen University, Guangzhou, China
}

\iclrfinalcopy 
\begin{document}

\maketitle

\begin{abstract}
The reasoning capabilities of Large Language Models (LLMs) are typically developed through the single-turn reinforcement learning, whereas real-world applications often involve multi-turn interactions with human feedback, leading to a potential mismatch between training and deployment conditions. In this work, we study whether multi-turn training with human feedback is necessary for reasoning tasks. We compare conventional single-turn training with three multi-turn strategies and reach contrary conclusions to previous research. We find that models trained in a single-turn setting generalize effectively to both single- and multi-turn evaluations, while models trained with multi-turn strategies exhibit a significant degradation in single-turn reasoning performance. These results suggest that for tasks with complete information, robust single-turn training remains more effective and reliable, as multi-turn training with basic feedback provides limited benefits and can even degrade reasoning capabilities.
\end{abstract}

\section{Introduction}
Recent developments of large language models (LLMs) enable them to solve mathematical problems and autonomously finish complex tasks composed of multiple steps of reasoning and information retrieval. Users do not need to interfere with the model's reasoning, decision-making, and execution process. They can focus only on the accuracy of the final output of the model. Reinforcement Learning (RL) has been a successful method in training the reasoning and decision-making abilities of LLMs. Verifiable answers, such as mathematical problem solutions, act as a reward model to supervise the training direction for reasoning-intensive tasks. An additional thinking section [\cite{guo2025deepseek,shao2024deepseekmath}) or Chain-of-Thought (CoT) [\cite{wei2022chain,kojima2022large}] allows the model to output its reasoning process tokens, essentially expanding the optimization/searching space of the problem solutions (\cite{snell2024scaling,muennighoff2025s1}). Furthermore, RL allows training of LLMs in real interactive environments (\cite{sheng2024hybridflow,qian2025toolrl,zhou2025sweet,wang2025ragen,lu2025arpo,zeng2025glm}]. 

The training of reasoning ability is conducted in a single-turn manner, where each rollout sample provides one final answer and ends with GRPO [\cite{feng2025group}], DAPO [\cite{yu2025dapo}], etc.. In practical applications, model inference can be performed in a multi-turn fashion. If a user receives a wrong answer, then the user gives feedback, such as the correctness of the answer, to the model, allowing it to reconsider the problem with the dialog history. Thus, there is an inconsistency between training and inference. However, it is straightforward to generalize single-turn training to multi-turn: Due to the alignment of the reward model and human preference in RL training, human feedback can be obtained from the reward model during training. Thus, multi-turn training with human feedback can be achieved by concatenating all contextual history to form the prompts of new turns.

Wang et. al. [\cite{wang2023mint}] evaluate the performance of open- and closed-source LLMs on multi-turn tasks, and they find that LLMs' performance between single- and multi-turn inference is inconsistent. Recent studies report that multi-turn training elicits multi-turn reasoning [\cite{liu2025let}] by providing basic feedback on correctness, termed lazy users by Wang et al. [\cite{wang2023mint}]. However, it is still insufficient to conclude that a multi-turn trained model performs better than a single-turn trained model: a multi-turn-trained model is evaluated with multi-turn inference (Succ@$K$) while a single-turn-trained model is evaluated with $K$ independent inferences (pass@$K$) [\cite{liu2025let}]. There is a natural performance gap between the two evaluations due to the uncertainty reduction of the answer in the multi-turn inference. Kumar et. al. [\cite{kumar2025score}] design a hand-crafted two-step optimization scheme to improve reasoning and reflection ability, but without decoupling the number of turns from the training scheme. Thus, it is still unclear how multi-turn human feedback/interaction influences the optimization of LLMs and whether multi-turn training is necessary or not. 

In this report, we clarify the concept of \textit{multi-step} and \textit{multi-turn}: human feedback divides turns; one turn contains multiple steps. We argue that multi-turn training is not necessary for reasoning tasks, e.g., solving mathematical problems: If the previous operation actions provide little information to the final-turn inference, then the model does not need to experience multi-turn samples during training. We design three multi-turn training strategies to cover possible optimization aspects in multi-turn training and evaluate the performance of single- and multi-turn trained models in both single- and multi-turn inference scenarios fairly. 

Our experiments on GSM8K show that the single-turn trained models perform well both in the single- and multi-turn inference; while multi-turn trained models perform well only in the multi-turn inference, but are insufficiently optimized for single-turn inference. 

\section{Reasoning ability enhancement with reinforcement learning}
\label{rlvr_algo}
Reinforcement learning (RL) has proven effective in enhancing the reasoning capabilities of large language models (LLMs), enabling cognitive behaviors such as self-reflection and error correction in complex tasks, including mathematical reasoning and code generation. Currently, the reasoning ability of LLMs is enhanced by training the models with reasoning-intensive tasks, e.g., solving mathematical problems, with reinforcement learning algorithms. A growing body of research focuses on developing simpler and more efficient training frameworks and reward mechanisms to further optimize model performance. For instance, PPO [\cite{schulman2017proximal}] improves reasoning while ensuring training stability through clipped policy updates; DPO [\cite{rafailov2023dpo}] eliminates the need for an explicit reward model by directly optimizing preference rankings based on human feedback; and GRPO [\cite{shao2024deepseekmath}] promotes exploration and reasoning via group-normalized reward signals. Additional methods such as GRPO dr. [\cite{liu2025understanding}] and DAPO [\cite{yu2025dapo}] further demonstrate that minimalist strategies—including decoupled clipping, unbiased optimization objectives, and simplified reward structures—can effectively enhance the reasoning abilities of LLMs. Open Reasoner Zero [\cite{hu2025openreasonerzero}] also shows similar effectiveness.

During the RL training, a verifiable reward function supervises the model to search optimized reasoning tokens before outputting the final answer, i.e. the 
Reinforcement Learning with Verifiable Rewards (RLVR) method. RLVR is approved to enhance the reasoning probability largely. In this report, all the experiments are done with GRPO with GSM8K dataset.
 
As an extension of PPO, GRPO eliminates the value function and estimates advantages through group-relative estimation. For each question $q$, GPRO samples a group of $G$ individual outputs $\{o_i\}^G_{i=1}$ from the behavior policy $\pi_{\theta_{old}}$. The advantage of the $i$-th output is then calculated through normalization relative to its output group's rewards $\{R_i\}^G_{i=1}$:
$$\hat{A}_{i,t}=\frac{r_i-\max(\{R_i\}_{i=1}^G)}{\operatorname{std}(\{R_i\}_{i=1}^G)}.$$

GRPO preserves PPO's clipped objective while imposing a supplementary KL-divergence constraint as a regularization mechanism and the training goal is to maximize:
$$\begin{aligned}
\mathcal{J}_{\mathrm{GRPO}}(\theta) & =\mathbb{E}_{(q,a)\sim\mathcal{D},\{o_i\}_{i=1}^G\sim\pi_{\theta_{\mathrm{old}}}(\cdot|q)} \\
 & \left[\frac{1}{G}\sum_{i=1}^G\frac{1}{|o_i|}\sum_{t=1}^{|o_i|}\left(\min\left(r_{i,t}(\theta)\hat{A}_{i,t},\mathrm{clip}\left(r_{i,t}(\theta),1-\varepsilon,1+\varepsilon\right)\hat{A}_{i,t}\right)-\beta D_{\mathrm{KL}}(\pi_\theta||\pi_{\mathrm{ref}})\right)\right],
\end{aligned}$$
where
$$r_{i,t}(\theta)=\frac{\pi_{\theta}(o_{i,t}\mid q,o_{i,<t})}{\pi_{\theta_{\mathrm{old}}}(o_{i,t}\mid q,o_{i,<t})}.$$

\section{Multi-turn reasoning}
We clarify the concept of multi-step and multi-turn in LLM operations for this report. Multi-step refers to the calling of external tools multiple times. Multi-turn reasoning refers to the scenario where the model autonomously does the reasoning job before outputting the final answer to human users. The significant difference between multi-step and multi-turn is whether the external information comes from the final reward model, corresponding to the situation that human users receive a final answer from the model and give feedback for improving the answer.

Multi-turn reasoning of an LLM agent can be denoted as an ordered list
\begin{equation}
\label{multi-turn}
[\underbrace{U_1, R_1, \overbrace{U_2, R_2}^{\mathrm{one~step}}, \cdots, U_n, R_n, F_1}_{\mathrm{one~turn}}, U_{n+1}, R_{n+1}, \cdots]   
\end{equation}

where $U$, $R$, and $F$ are external messages (such as tool responses), model response, and human feedback, respectively. Each step is defined as a round of new input tokens $U$ followed by a model response, and a turn ends with the final result response and human feedback. In each step, the response is generated by prompting the model with all the history tokens. One turn can be composed of one step, or multiple steps if the model needs to call external tools to gather mandatory information. $K$-turn reasoning means that the maximum turn allowed is $K$.

$K$-turn reasoning is similar to a pass@$K$ reasoning where the model is called $K$-times independently. The advantage of the $K$-turn reasoning over pass@$K$ is that the reasoning model becomes aware of the previous wrong answers, and there is a reduction of the uncertainty of the correct answer. Furthermore, the model may implicitly take advantage of the previous reasoning turns to enhance the correctness of the final-turn reasoning. Multi-turn reasoning is a common scenario, ranging from human users querying the model for a satisfied answer by chatting, to the model being applied to deal with complex tasks. Whether a question should be solved in a multi-turn manner may be decided by the information completeness, see Sec.\ref{apped:complete}.

\section{Single-turn and multi-turn reasoning ability}
\label{sec:methedology}
The multi-turn reasoning scenario is common in real applications, but current open source reasoning models, such as Deepseep-R1 [\cite{guo2025deepseek}], Qwen series [\cite{yang2025qwen3}], etc., are trained in a single-turn style as introduced in Sec.~\ref{rlvr_algo}. Natural questions that arise:

\begin{itemize}
  \item Can single-turn reasoning abilities extend to multi-turn reasoning scenarios? 
  \item Should human feedback be introduced into reasoning training to align the training and inference application?
\end{itemize}

To answer the above questions, we design a training framework which is also implemented by an open source training framework \textit{VeRL} (\textit{Interaction} module) and Liu et. al. [\cite{liu2025let}]: the final answer is verified by the reward model and if the response is not correct, then negative feedback is given to prompt the model to continue generation. Since the training goal is to align with human requirements, the reward model feedback can substitute human feedback in the training rollout. In our experiment on solving GSM8K [\cite{cobbe2021gsm8k}] math problems, we use the following sentence as human feedback $F_i$ when $R_i$ is wrong in (\ref{multi-turn}): \textit{Your response is incorrect, or your answer is not given in the correct form. You need to reflect on your answer and try again}.

We report experimental results of single- and multi-turn training model performance with three different training strategies on GSM8K: model Update at All responses with Consistent Reward (UACR), model Update at Last response with Consistent Reward (ULCR), and model Update at All responses with Decay Reward (UADR). The difference among strategies lies in the model gradient update and reward design:
\begin{itemize}
  \item UACR: The model updates its parameters at all the tokens located in responses (all $R_x$ in Eq.~(\ref{multi-turn})) and if the final response of the last turn gives a correct answer, then the reward is $1$;
  \item ULCR: The model updates its parameters only at the tokens located in the final response of the last turn. If the final response of the last turn gives a correct answer, then the reward is $1$;
  \item UADR: The model updates its parameters at all the tokens located in responses (same as UACR). However, the reward is progressively discounted according to the number of turns taken to reach the correct answer as follows:
\begin{equation}
\label{reward-decay}
    r = \frac{1}{\log_{2}(t + 1)}, t = \text{steps to reach the correct answer} 
\end{equation}

\end{itemize}
The model parameters should only be updated at the response tokens that are generated by the model itself. UACR is a plain strategy that the model updates at the model responses, and the final reward simply indicates whether the model rollout is correct or not. ULCR is under the fact that all responses except the last one provide wrong answers in a positive sample with a reward $1$. To eliminate the inconsistency, only the last response is used to calculate advantages and update model parameters. UADR incorporates the goals that shorten the inference turns and reduce their reward: more responses, more wrong answers.

For single-turn training, all three above strategies are equivalent. The last response is the only response of single-turn inference, and the decay reward is always $1$ for a sample containing a correct answer since the number of steps is $t=1$.

The result models trained by the above strategies are evaluated by both inference scenarios: $K$-turn and Pass@$K$. For $K$-turn inference, the model continuously generated responses with a maximum $K$-turn. If the model outputs a correct answer, then the evaluation on the particular sample is accurate. The $K$-turn evaluation mimics real applications where users only focus on the final results but not the complicated inner operations of the agent models. For Pass@$K$, the model generates $K$ answers independently, and each answer is generated within a single turn. If the model generates at least one correct answer among $K$ answers, then the evaluation is accurate. Pass@$K$ evaluates the model's single-turn performance independently $K$ times with roughly comparable metric values to that of $K$-turn: In both scenarios, the model has $K$ chances to output a correct answer. The difference is that the generation of new responses may take advantage of the old wrong responses in $K$-turn evaluation.

The comparison of metrics should be fair: Pass@$K$ and $K$-turn accuracy should not be conflated, even though the two scenarios share similar computational complexity. A natural performance gap exists between Pass@$K$ and $K$-turn accuracy due to the reduction in answer uncertainty during $K$-turn inference. 

\section{Experimental results}
We train the Qwen2.5-3B-Instruct model [\cite{qwen2025qwen25technicalreport}] with the above three strategies using GRPO on the GSM8K dataset. We use the VeRL [\cite{sheng2024hybridflow}] framework to conduct the follow experiments. The training lasts for $5$ epochs, and after each training step, we validate the model on test samples. In all the experiments, we evaluate the model with $8$-turn and Pass@$8$ accuracy. To obtain results, we set the temperature parameter $0$ in the validation stage.

\subsection{UACR and ULCR: Single-turn Performance Gap between Single- and Multi-turn Training}
\label{SEC:UACR_ULCR}
\begin{figure}
    \centering
    \includegraphics[width=1\linewidth]{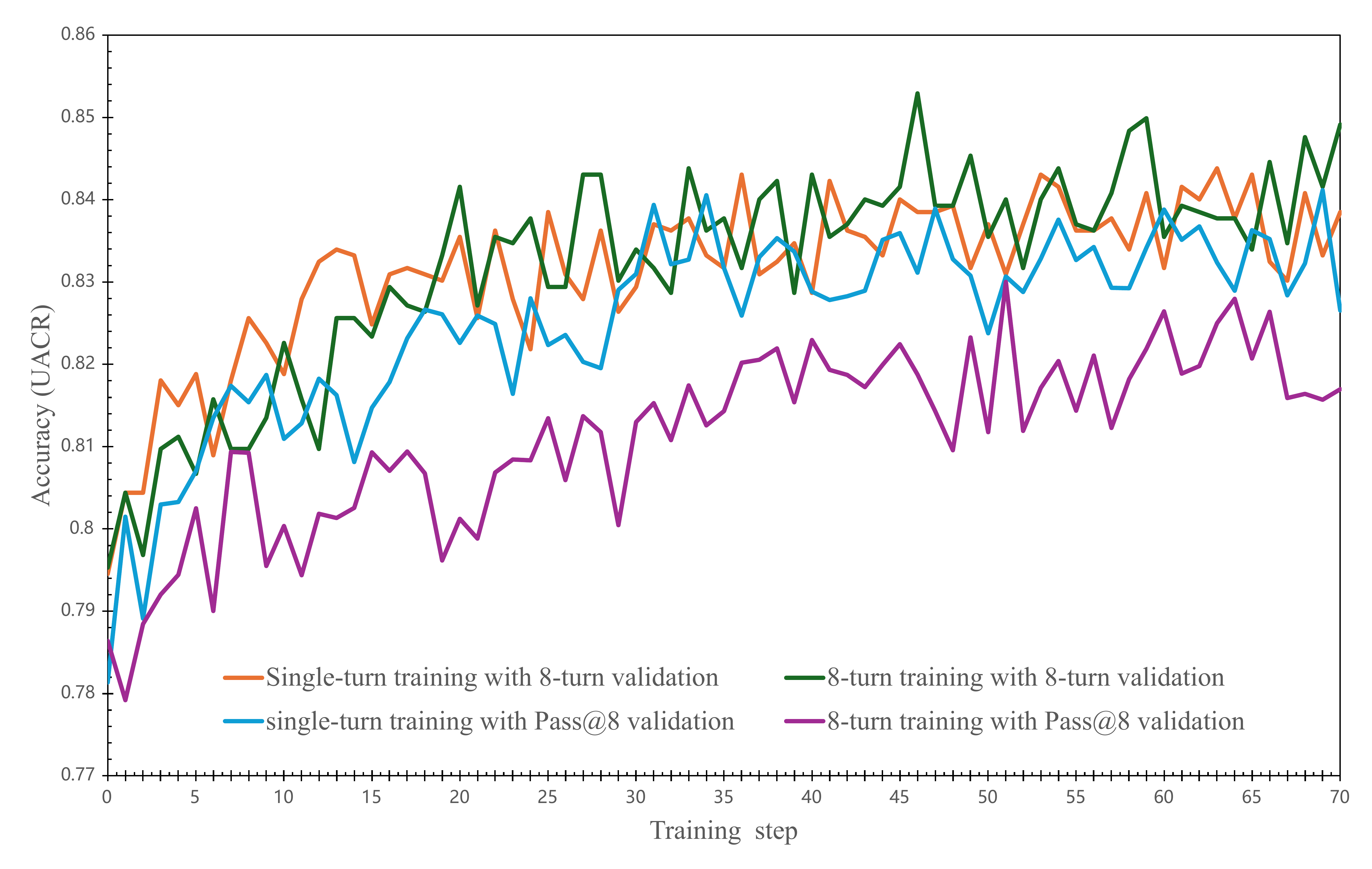}
    \caption{UACR experimental results: both single- and multi- turn trained models are evaluated with Pass@$8$ and $8$-turn inferences. Multi-turn trained model shows an inferior performance on single-turn inference and no obvious performance advantage on multi-turn inference}
    \label{fig:UACR}
\end{figure}

\begin{figure}
    \centering
    \includegraphics[width=1\linewidth]{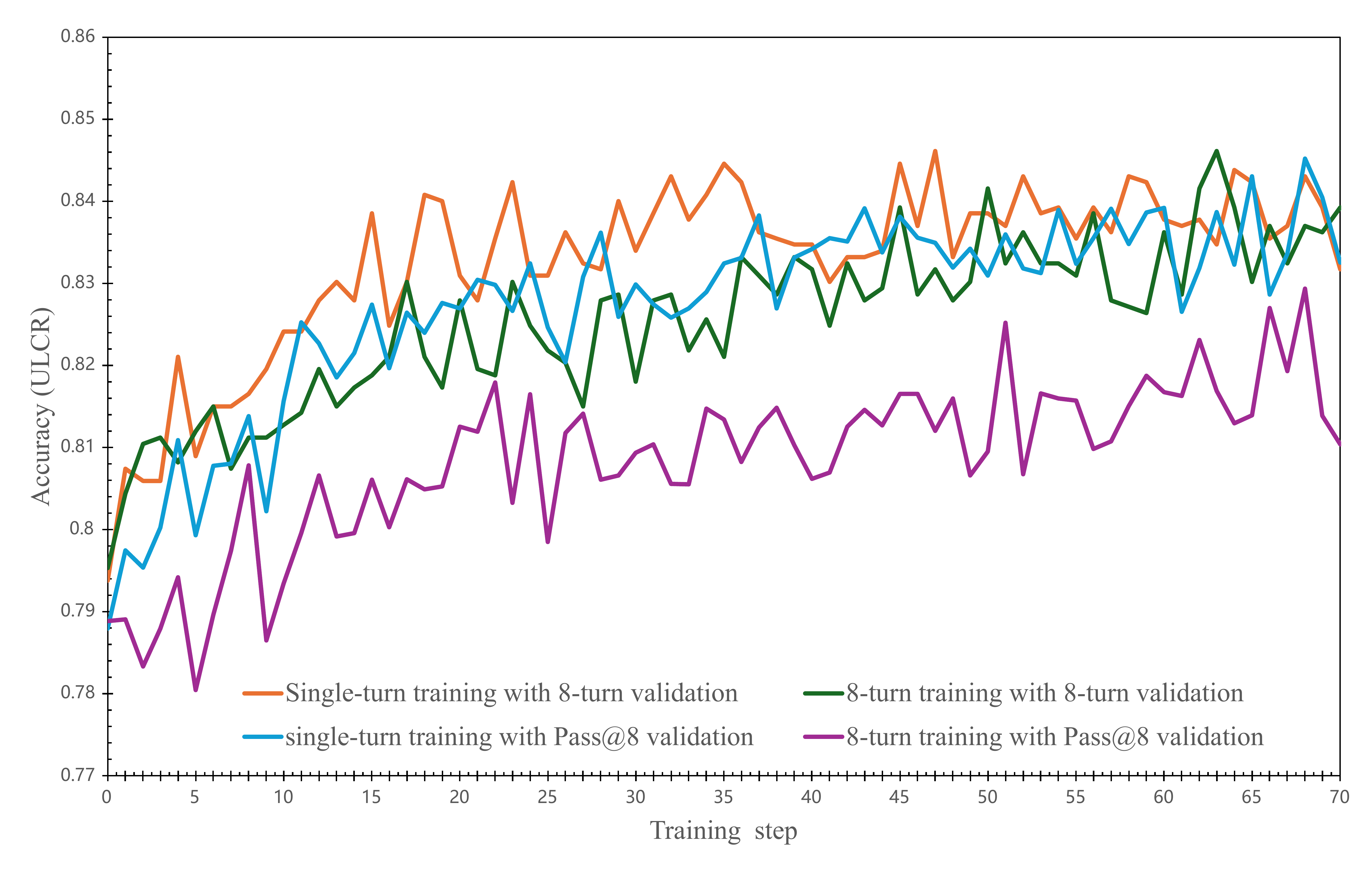}
    \caption{ULCR experimental results: similar to that of UACR}
    \label{fig:ULCR}
\end{figure}

We find an obvious performance gap on test samples between the single- and multi-turn training models on single-turn inference as indicated in Fig.~\ref{fig:UACR} and Fig.~\ref{fig:ULCR}, no matter the optimization strategy. 

Both single- and multi-turn trained models perform similarly on $8$-turn inferences. The accuracy of the single-turn trained model with Pass@$8$ validation is also similar to that of the $8$-turn inference, which indicates that previous wrong reasoning responses add subtle value to the final correct reasoning response for both single- and multi-turn trained models during multi-turn inference. 

However, there is an obvious weaker performance for the multi-turn trained model on single-turn inference. Multi-turn training is not an optimized way to argue a model's single reasoning ability, but single-turn training can augment a model's performance on both single- and multi-turn inference.

\subsection{UADR: Reward the Model Solving with Fewer Turns}
\label{SEC:UADR}
\begin{figure}
    \centering
    \includegraphics[width=1\linewidth]{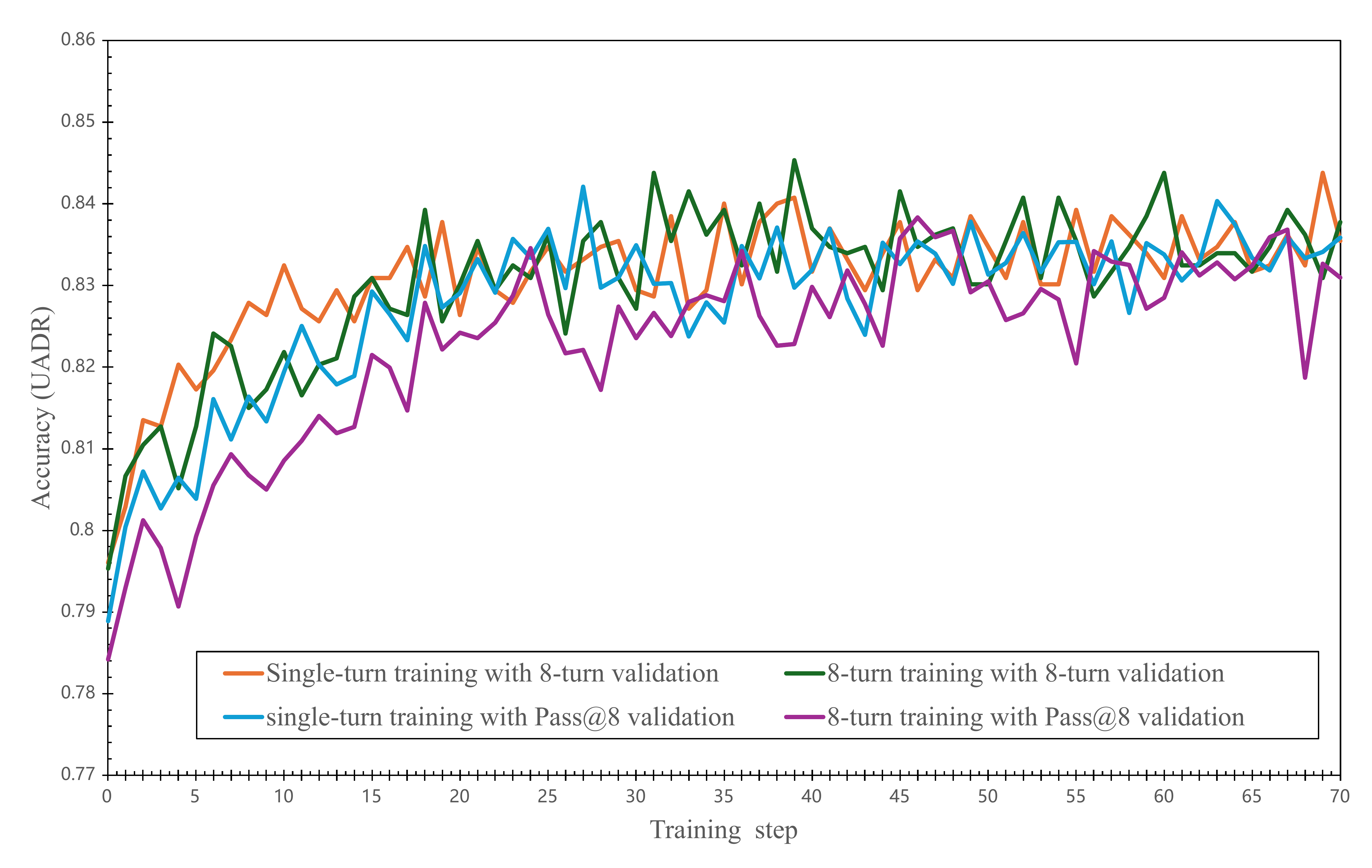}
    \caption{UADR experimental results: similar to that of UACR, but there is an improvement of the multi-turn trained model performing on single-turn inference}
    \label{fig:UADR}
\end{figure}

UADR training strategy encourages the model to solve the problem with fewer turns. The reward function, as indicated in Eq.~(\ref{reward-decay}), is monotonically decreasing in the number of turns. In the UADR strategy, we do not see an obvious performance\footnote{For UADR, the accuracy is not a reward and the value of accuracy is larger than the value of reward. For all the experiments, we unify the calculation method of accuracy on test samples as introduced in Sec.~\ref{sec:methedology}} Differences in test samples among the scenarios as indicated in Fig.~\ref{fig:UADR}. While UACR and ULCR only care about the correctness of the final answer, omitting the number of turns, single- and multi-turn UADR trainings are with no distinct border. A well-trained multi-turn UADR model maximizes its chance of solving the problem within fewer turns to obtain a larger reward. 

\subsection{Compare among three strategies}
We roughly cover possible training strategies with UACR, ULCR, and UADR. UACR and ULCR mimic different updating strategies that treat correct and incorrect answers differently. UADR represents a class of decay reward designs due to the weak correlation between the final correct answer and previous wrong tries. In addition, decay rewards encourage the model to solve the problem with minimum consumption of computational resources. 

\begin{figure}
    \centering
    \includegraphics[width=1\linewidth]{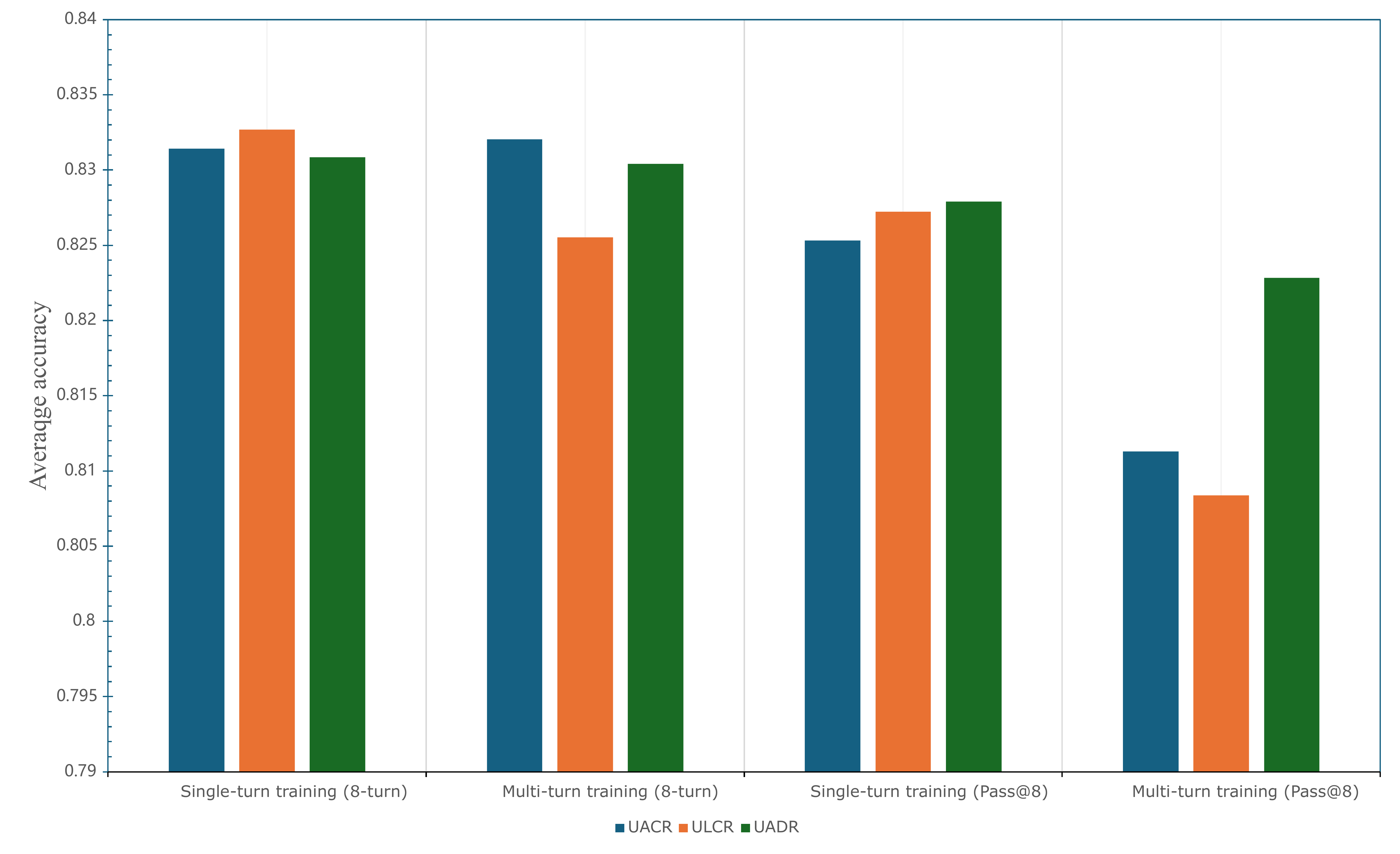}
    \caption{An overview of the three training strategies by averaging over steps}
    \label{fig:avg_three_algo}
\end{figure}

We average the accuracy over all steps in Fig.~\ref{fig:avg_three_algo}. For single-turn training, UACR, ULCR, and UADR are equivalent since there is only one final response. As indicated in Fig.~\ref{fig:avg_three_algo}, the performances are equivalent with small variations\footnote{The training parameter \textit{gpu\_memory\_utilization} is set as $0.2$, $0.2$, and $0.5$ for UACR, UADR, and ULCR due to the training complexity, respectively. Theoretically, this parameter does not influence the accuracy performance. We regard the accuracy difference as noisy variations. On total $1319$ samples in the validation set, and there in average $1.69$ to $2.52$ more wrong samples for UACR compared to ULCR}. 

Multi-turn training UACR performs better than ULCR in both Pass@8 and $8$-turns as indicated in Fig.~\ref{fig:avg_three_algo}, which is counterintuitive since the model reinforces its policy with previous wrong responses with a positive rollout in UACR.

All experimental results with three strategies show that multi-turn trained models have no obvious advantage compared to single-turn trained models on multi-turn inference ($K$-turn). Furthermore, multi-turn trained models decay on single-turn inference (Pass@$K$) performance. The performance decay in UADR is less compared to that in UACR and ULCR as UADR encourages model to solve the task within fewer turns or one turn.

\section{Conclusions}
There is an inconsistency between single-turn reasoning argumentation and multi-turn inference with human feedback of LLMs. Whether advantages can be obtained by introducing human feedback into the training process is insufficiently studied. We designed three multi-turn optimization strategies to evaluate this issue. Contrary to current knowledge, we show that multi-turn training with basic human feedback helps little on multi-turn LLM reasoning and even reduces its single-turn reasoning ability.

\bibliography{iclr2025_conference}

\begin{thebibliography}{24}
\providecommand{\natexlab}[1]{#1}
\providecommand{\url}[1]{\texttt{#1}}
\expandafter\ifx\csname urlstyle\endcsname\relax
  \providecommand{\doi}[1]{doi: #1}\else
  \providecommand{\doi}{doi: \begingroup \urlstyle{rm}\Url}\fi

\bibitem[Cobbe et~al.(2021)Cobbe, Kosaraju, Bavarian, Chen, Jun, Kaiser, Plappert, Tworek, Hilton, Nakano, Hesse, and Schulman]{cobbe2021gsm8k}
Karl Cobbe, Vineet Kosaraju, Mohammad Bavarian, Mark Chen, Heewoo Jun, Lukasz Kaiser, Matthias Plappert, Jerry Tworek, Jacob Hilton, Reiichiro Nakano, Christopher Hesse, and John Schulman.
\newblock Training verifiers to solve math word problems.
\newblock \emph{arXiv preprint arXiv:2110.14168}, 2021.

\bibitem[Feng et~al.(2025)Feng, Xue, Liu, and An]{feng2025group}
Lang Feng, Zhenghai Xue, Tingcong Liu, and Bo~An.
\newblock Group-in-group policy optimization for llm agent training.
\newblock \emph{arXiv preprint arXiv:2505.10978}, 2025.

\bibitem[Guo et~al.(2025)Guo, Yang, Zhang, Song, Zhang, Xu, Zhu, Ma, Wang, Bi, et~al.]{guo2025deepseek}
Daya Guo, Dejian Yang, Haowei Zhang, Junxiao Song, Ruoyu Zhang, Runxin Xu, Qihao Zhu, Shirong Ma, Peiyi Wang, Xiao Bi, et~al.
\newblock Deepseek-r1: Incentivizing reasoning capability in llms via reinforcement learning.
\newblock \emph{arXiv preprint arXiv:2501.12948}, 2025.

\bibitem[Hu et~al.(2025)Hu, Zhang, Han, Jiang, Zhang, and Shum]{hu2025openreasonerzero}
Jingcheng Hu, Yinmin Zhang, Qi~Han, Daxin Jiang, Xiangyu Zhang, and Heung-Yeung Shum.
\newblock Open-reasoner-zero: An open source approach to scaling up reinforcement learning on the base model.
\newblock \emph{arXiv preprint arXiv:2503.24290}, 2025.

\bibitem[Kojima et~al.(2022)Kojima, Gu, Reid, Matsuo, and Iwasawa]{kojima2022large}
Takeshi Kojima, Shixiang~Shane Gu, Machel Reid, Yutaka Matsuo, and Yusuke Iwasawa.
\newblock Large language models are zero-shot reasoners.
\newblock \emph{Advances in neural information processing systems}, 35:\penalty0 22199--22213, 2022.

\bibitem[Kumar et~al.(2025)Kumar, Zhuang, Agarwal, Su, Co{-}Reyes, Singh, Baumli, Iqbal, Bishop, Roelofs, Zhang, McKinney, Shrivastava, Paduraru, Tucker, Precup, Behbahani, and Faust]{kumar2025score}
Aviral Kumar, Vincent Zhuang, Rishabh Agarwal, Yi~Su, John~D. Co{-}Reyes, Avi Singh, Kate Baumli, Shariq Iqbal, Colton Bishop, Rebecca Roelofs, Lei~M. Zhang, Kay McKinney, Disha Shrivastava, Cosmin Paduraru, George Tucker, Doina Precup, Feryal M.~P. Behbahani, and Aleksandra Faust.
\newblock Training language models to self-correct via reinforcement learning.
\newblock In \emph{The Thirteenth International Conference on Learning Representations, {ICLR} 2025, Singapore, April 24-28, 2025}, 2025.
\newblock URL \url{https://openreview.net/forum?id=CjwERcAU7w}.

\bibitem[Liu et~al.()Liu, Wang, Li, Xu, Lu, Liu, Sil, and Li]{liu2025let}
Licheng Liu, Zihan Wang, Linjie Li, Chenwei Xu, Yiping Lu, Han Liu, Avirup Sil, and Manling Li.
\newblock Let’s try again: Eliciting multi-turn reasoning in language models via simplistic feedback.
\newblock In \emph{2nd AI for Math Workshop@ ICML 2025}.

\bibitem[Liu et~al.(2025)Liu, Chen, Li, Qi, Pang, Du, Lee, and Lin]{liu2025understanding}
Zichen Liu, Changyu Chen, Wenjun Li, Penghui Qi, Tianyu Pang, Chao Du, Wee~Sun Lee, and Min Lin.
\newblock Understanding r1-zero-like training: A critical perspective.
\newblock \emph{arXiv preprint arXiv:2503.20783}, 2025.

\bibitem[Lu et~al.(2025)Lu, Zhong, Liu, Fu, and Jia]{lu2025arpo}
Fanbin Lu, Zhisheng Zhong, Shu Liu, Chi-Wing Fu, and Jiaya Jia.
\newblock Arpo: End-to-end policy optimization for gui agents with experience replay.
\newblock \emph{arXiv preprint arXiv:2505.16282}, 2025.

\bibitem[Muennighoff et~al.(2025)Muennighoff, Yang, Shi, Li, Fei-Fei, Hajishirzi, Zettlemoyer, Liang, Cand{\`e}s, and Hashimoto]{muennighoff2025s1}
Niklas Muennighoff, Zitong Yang, Weijia Shi, Xiang~Lisa Li, Li~Fei-Fei, Hannaneh Hajishirzi, Luke Zettlemoyer, Percy Liang, Emmanuel Cand{\`e}s, and Tatsunori Hashimoto.
\newblock s1: Simple test-time scaling.
\newblock \emph{arXiv preprint arXiv:2501.19393}, 2025.

\bibitem[Qian et~al.(2025)Qian, Acikgoz, He, Wang, Chen, Hakkani-T{\"u}r, Tur, and Ji]{qian2025toolrl}
Cheng Qian, Emre~Can Acikgoz, Qi~He, Hongru Wang, Xiusi Chen, Dilek Hakkani-T{\"u}r, Gokhan Tur, and Heng Ji.
\newblock Toolrl: Reward is all tool learning needs.
\newblock \emph{arXiv preprint arXiv:2504.13958}, 2025.

\bibitem[Qwen et~al.(2025)Qwen, :, Yang, Yang, Zhang, Hui, Zheng, Yu, Li, Liu, Huang, Wei, Lin, Yang, Tu, Zhang, Yang, Yang, Zhou, Lin, Dang, Lu, Bao, Yang, Yu, Li, Xue, Zhang, Zhu, Men, Lin, Li, Tang, Xia, Ren, Ren, Fan, Su, Zhang, Wan, Liu, Cui, Zhang, and Qiu]{qwen2025qwen25technicalreport}
Qwen, :, An~Yang, Baosong Yang, Beichen Zhang, Binyuan Hui, Bo~Zheng, Bowen Yu, Chengyuan Li, Dayiheng Liu, Fei Huang, Haoran Wei, Huan Lin, Jian Yang, Jianhong Tu, Jianwei Zhang, Jianxin Yang, Jiaxi Yang, Jingren Zhou, Junyang Lin, Kai Dang, Keming Lu, Keqin Bao, Kexin Yang, Le~Yu, Mei Li, Mingfeng Xue, Pei Zhang, Qin Zhu, Rui Men, Runji Lin, Tianhao Li, Tianyi Tang, Tingyu Xia, Xingzhang Ren, Xuancheng Ren, Yang Fan, Yang Su, Yichang Zhang, Yu~Wan, Yuqiong Liu, Zeyu Cui, Zhenru Zhang, and Zihan Qiu.
\newblock Qwen2.5 technical report, 2025.
\newblock URL \url{https://arxiv.org/abs/2412.15115}.

\bibitem[Rafailov et~al.(2023)Rafailov, Sharma, Mitchell, Manning, Ermon, and Finn]{rafailov2023dpo}
Rafael Rafailov, Archit Sharma, Eric Mitchell, Christopher~D. Manning, Stefano Ermon, and Chelsea Finn.
\newblock Direct preference optimization: Your language model is secretly a reward model.
\newblock In \emph{Advances in Neural Information Processing Systems 36: Annual Conference on Neural Information Processing Systems 2023, NeurIPS 2023, New Orleans, LA, USA, December 10 - 16, 2023}, 2023.

\bibitem[Schulman et~al.(2017)Schulman, Wolski, Dhariwal, Radford, and Klimov]{schulman2017proximal}
John Schulman, Filip Wolski, Prafulla Dhariwal, Alec Radford, and Oleg Klimov.
\newblock Proximal policy optimization algorithms.
\newblock \emph{arXiv preprint arXiv:1707.06347}, 2017.

\bibitem[Shao et~al.(2024)Shao, Wang, Zhu, Xu, Song, Bi, Zhang, Zhang, Li, Wu, et~al.]{shao2024deepseekmath}
Zhihong Shao, Peiyi Wang, Qihao Zhu, Runxin Xu, Junxiao Song, Xiao Bi, Haowei Zhang, Mingchuan Zhang, YK~Li, Yang Wu, et~al.
\newblock Deepseekmath: Pushing the limits of mathematical reasoning in open language models.
\newblock \emph{arXiv preprint arXiv:2402.03300}, 2024.

\bibitem[Sheng et~al.(2024)Sheng, Zhang, Ye, Wu, Zhang, Zhang, Peng, Lin, and Wu]{sheng2024hybridflow}
Guangming Sheng, Chi Zhang, Zilingfeng Ye, Xibin Wu, Wang Zhang, Ru~Zhang, Yanghua Peng, Haibin Lin, and Chuan Wu.
\newblock Hybridflow: A flexible and efficient rlhf framework.
\newblock \emph{arXiv preprint arXiv: 2409.19256}, 2024.

\bibitem[Snell et~al.(2024)Snell, Lee, Xu, and Kumar]{snell2024scaling}
Charlie Snell, Jaehoon Lee, Kelvin Xu, and Aviral Kumar.
\newblock Scaling llm test-time compute optimally can be more effective than scaling model parameters.
\newblock \emph{arXiv preprint arXiv:2408.03314}, 2024.

\bibitem[Wang et~al.(2023)Wang, Wang, Liu, Chen, Yuan, Peng, and Ji]{wang2023mint}
Xingyao Wang, Zihan Wang, Jiateng Liu, Yangyi Chen, Lifan Yuan, Hao Peng, and Heng Ji.
\newblock Mint: Evaluating llms in multi-turn interaction with tools and language feedback.
\newblock \emph{arXiv preprint arXiv:2309.10691}, 2023.

\bibitem[Wang et~al.(2025)Wang, Wang, Wang, Zhang, Li, Yang, Jin, Yu, Nguyen, Liu, et~al.]{wang2025ragen}
Zihan Wang, Kangrui Wang, Qineng Wang, Pingyue Zhang, Linjie Li, Zhengyuan Yang, Xing Jin, Kefan Yu, Minh~Nhat Nguyen, Licheng Liu, et~al.
\newblock Ragen: Understanding self-evolution in llm agents via multi-turn reinforcement learning.
\newblock \emph{arXiv preprint arXiv:2504.20073}, 2025.

\bibitem[Wei et~al.(2022)Wei, Wang, Schuurmans, Bosma, Xia, Chi, Le, Zhou, et~al.]{wei2022chain}
Jason Wei, Xuezhi Wang, Dale Schuurmans, Maarten Bosma, Fei Xia, Ed~Chi, Quoc~V Le, Denny Zhou, et~al.
\newblock Chain-of-thought prompting elicits reasoning in large language models.
\newblock \emph{Advances in neural information processing systems}, 35:\penalty0 24824--24837, 2022.

\bibitem[Yang et~al.(2025)Yang, Li, Yang, Zhang, Hui, Zheng, Yu, Gao, Huang, Lv, et~al.]{yang2025qwen3}
An~Yang, Anfeng Li, Baosong Yang, Beichen Zhang, Binyuan Hui, Bo~Zheng, Bowen Yu, Chang Gao, Chengen Huang, Chenxu Lv, et~al.
\newblock Qwen3 technical report.
\newblock \emph{arXiv preprint arXiv:2505.09388}, 2025.

\bibitem[Yu et~al.(2025)Yu, Zhang, Zhu, Yuan, Zuo, Yue, Dai, Fan, Liu, Liu, et~al.]{yu2025dapo}
Qiying Yu, Zheng Zhang, Ruofei Zhu, Yufeng Yuan, Xiaochen Zuo, Yu~Yue, Weinan Dai, Tiantian Fan, Gaohong Liu, Lingjun Liu, et~al.
\newblock Dapo: An open-source llm reinforcement learning system at scale.
\newblock \emph{arXiv preprint arXiv:2503.14476}, 2025.

\bibitem[Zeng et~al.(2025)Zeng, Lv, Zheng, Hou, Chen, Xie, Wang, Yin, Zeng, Zhang, et~al.]{zeng2025glm}
Aohan Zeng, Xin Lv, Qinkai Zheng, Zhenyu Hou, Bin Chen, Chengxing Xie, Cunxiang Wang, Da~Yin, Hao Zeng, Jiajie Zhang, et~al.
\newblock Glm-4.5: Agentic, reasoning, and coding (arc) foundation models.
\newblock \emph{arXiv preprint arXiv:2508.06471}, 2025.

\bibitem[Zhou et~al.(2025)Zhou, Jiang, Tian, Weston, Levine, Sukhbaatar, and Li]{zhou2025sweet}
Yifei Zhou, Song Jiang, Yuandong Tian, Jason Weston, Sergey Levine, Sainbayar Sukhbaatar, and Xian Li.
\newblock Sweet-rl: Training multi-turn llm agents on collaborative reasoning tasks.
\newblock \emph{arXiv preprint arXiv:2503.15478}, 2025.

\end{thebibliography}
\bibliographystyle{iclr2025_conference}

\appendix
\section{Complete and incomplete information: a rough categorization of tasks}
\label{apped:complete}
We may roughly categorize multi-turn reasoning tasks into two kinds: the complete and incomplete information tasks. A complete information task means that the needed information for solving the task within one turn is complete, and the output of a turn depends little on previous turns. The task is solved independently and repeatedly among turns. Following Eq.~(\ref{multi-turn}), a task is complete information if
\begin{equation}
    H(X|R_n, F_1) \approx H(X|U_1, R_1, U_2, R_2, \cdots, U_n, R_n, F_1)
\end{equation}
Where $H(X)$ is the entropy of the answer $X$. For the second round inference of a model, the uncertainty deduction comes only from human feedback on the final results of the previous turn, e.g., a hint about the correctness of $R_n$ excluding a wrong answer. 

For incomplete information tasks, the previous operation process $U_1, R_1, \cdots, U_n, R_n$ is fairly necessary for following turns:
\begin{equation}
    H(X|R_n, F_1) \ll H(X|U_1, R_1, U_2, R_2, \cdots, U_n, R_n, F_1).
\end{equation} 

We argue that most of the multi-turn inference is complete information. A typical example is solving mathematical problems. The problem solver responds with the reasoning process and the answer. The verifier gives feedback on whether the answer is correct or not. If the answer is wrong, the solver can redo the calculation. The previous incorrect calculation process is not necessary to deduce a correct calculation. For an incomplete information task, each turn solves an intermediate goal for the final result.  Multi-turn training may not be necessary for complete information tasks: If the previous operation actions provide little information to the final-turn inference, then the model does not need to experience multi-turn samples during training. 

\end{document}